# Semantic Chains: From Reachable Facts to Justified Answers

**Misha Andreev**

andreevmichael90@gmail.com

## Introduction

Context provides information for sustaining a dialogue, identifying the causes and consequences of events, and determining the conditions needed to understand what is happening. The meaning of a statement in context depends on its own content, its relations to other statements, and the information surrounding it, including how participants are designated, temporal information, message structure, and other conditions relevant to interpretation.*

This work develops a mathematical model of how such information is connected. The central research question is how a question relates to the logical justification of its answer and under what conditions that justification can be presented as a sequence of connected facts.** The initial hypotheses and premises emerged from work on datasets for evaluating answers generated by large language models. Two considerations are important in such evaluations: whether the facts found justify the answer and whether its wording preserves the relation asked about.

Particular attention is given to the geometry of relations between facts: how facts are organized in context, which of their relations form the justification of an answer to a particular question, and in what order the facts of that justification can be presented as a chain. In the conceptual graph model, an answer graph must preserve the roles, relations, and known values specified in the query graph and supply values for the participants sought.***

A search starting from the objects in the question discovers facts in the contract that are connected to them. From these facts, the justification uses those through which the requested relation is expressed. To construct a chain, we must establish which transformations preserve that relation in the content of the links.

We introduce the conditions and basic concepts of the model, then state and prove results on the structure of a full justification and the transformations that preserve it.

---

* Halliday & Hasan (1976); Morris & Hirst (1991); Hobbs (1979); Clark & Brennan (1991); Clark & Wilkes-Gibbs (1986); van Dijk & Kintsch (1983); Kamp, van Genabith & Reyle (2011).

** Roberts (2012); Graesser & Franklin (1990).

*** Sowa (1976); Khashabi et al. (2018).

## Definitions and Model Conditions

## Contract Structure

T1. The contract comprises the main context and its frame, which the participants draw on to understand the question and justify the answer. The main context contains the messages and statements under consideration. The participants share an interpretation of the accepted facts and their designations.

T2. The frame includes the language and interpretive conventions, the title, headings and structure of the source, participant roles, dates, times, and other information needed to interpret the content.

T3. The contract must supply all the initial information needed to answer the question and the grounds for interpreting that information. Supplementary information supplied separately from the main context forms part of the contract. Data can belong to the contract even when they occur outside a message or dialogue. Completeness of the initial data is assessed relative to the question and the contract as a whole. For example, counting events over a period requires establishing which events fall within it, distinguishing separate events from repeated mentions of the same event, and confirming that the count covers all of them. Filtering messages by date alone cannot establish completeness.

T4. Content is considered in lexical form under an agreed linguistic interpretation. This applies to continuous prose, lists, dialogues, and other structured sources. Multilingual material and formal symbols require an established translation of their relevant content into this form. Numerical representations of dates, times, and other data are admissible when their interpretation is unambiguous.

T5. Source structure is taken into account to the extent that it expresses established relations. A heading may place several items within a single episode. Dialogue markup may identify speakers, participant roles, and the order of messages. The order of messages and the order of the events they describe can differ. Similarly, the date of a message alone does not establish the date of the event it reports.

T6. Metadata may first be put into words or processed separately by a technical component. In the latter case, the data must remain verifiably associated with the original messages and events, and the result used must admit an unambiguous lexical expression. The entire source need not be transcribed.

T7. An object has a designation in the question or contract and is treated as indivisible in the selected analysis. In particular, it contains no further objects within that analysis.

T8. A fact is an accepted statement about objects, their properties, relations, or events.

T8.1. The content of a fact includes the relations between objects and their roles as specified by the lexical construction.

T8.2. Every fact used contains at least one object with a verifiable designation. If the object is not named directly, its designation is established through a confirmed reference.

T8.3. Preserving transformations use references to facts that admit full content-preserving expansion under P2. Resolving the target alone is insufficient: the referring fact must justify the content of its target. Substituting a designation under T9.2.2 does not automatically incorporate the asserted content of the target.

T8.4. The logical fact is the unit of the contract and the vertex of the answer graph. Full, abbreviated, and synonymous expressions that designate the same logical fact with all its established content in this contract form the family of lexical representations of that fact. If the fact is included in the selected justification, every expression in this family corresponds to the same vertex of the answer graph. A logical fact is identified by its content: the identity of events and objects, relations, participant roles, characteristics, conditions, and their scope. The locations of mentions are recorded as source addresses rather than separate logical facts.

T8.5. Extracting nested facts preserves the conditions, assumptions, and attribution of the utterance to a participant expressed in the parent construction. Extraction must not turn conditional or hypothetical content into an unconditional assertion.

T9. Structural connections in the question and contract take the form of inclusion and reference.

T9.1. In an inclusion, a fact contains an object or a subfact.

T9.2. In a reference, an expression in a question or fact designates a particular object or fact. It may be a name, pronoun, description, or other referring expression. We distinguish the expression, its target, and the evidence that establishes their correspondence.

T9.2.1. The correspondence may be established directly in the fact containing the expression or confirmed by other facts in the contract. In the latter case, the supporting facts form a finite set, and each step of reference resolution is checked.

T9.2.2. If the contract unambiguously establishes that two expressions designate the same object or fact, one may replace the other in the text under consideration, provided its content, participants, roles, and conditions are preserved. The correspondence holds in the given context and does not imply general synonymy. Its supporting evidence remains available for verification. Replacing a designation does not automatically add separate assertions from that evidence to the text.

T9.3. Referring to an object does not automatically include all facts about that object.

T9.4. Full preservation. A⩾B means that A includes the entire asserted content of B, preserving the identity of events and objects, roles, conditions, and designations. The relation is reflexive and transitive. Coincidence of logical consequences alone is insufficient. Strict edges connect distinct logical facts. An address that refers back to the same logical fact creates no additional vertex.

T10. A parent fact includes all subfacts that can be distinguished within its content while preserving the meaning, object designations, roles, and conditions of the respective assertions. A single predefined segmentation of the text does not limit the choice of subfacts.

T10.1. A parent fact expresses a relation between the logical facts nested within it. Each nested fact is a subfact of that parent. One subfact may belong to more than one parent fact.

T10.2. The reflexive expressions F⩾F and F⩽F indicate identity and do not constitute strict nesting. For example, “David went to the library yesterday” is one fact with all the established attributes of the event. Expressions such as “David went” and “David went yesterday,” when used in the contract to designate this same fact, belong to its family of lexical representations. They are neither separate subfacts nor separate vertices. Attributes omitted from an expression remain part of the full fact it designates.

T11. Every object of the question and contract is designated in at least one accepted fact.

T12. The question and contract contain finitely many facts.

T13. The accepted facts of the question and contract are mutually consistent.

## Question and Answer Structure

K1. Objects, facts, inclusions, and references in the question and contract are constructed according to the same rules.

K1.1. The question specifies what information is sought and refers to at least one object. Facts and objects may be named directly or identified by references confirmed by the contract. The facts, objects, and connections indicated in the question must correspond to the contract with their meanings and roles preserved. The facts specified by the question and the selected facts supporting the full answer are considered within the same contract.

K1.2. If a question can be divided into independent questions, each can be answered separately under the given contract without establishing a connection between them.

K2. The model considers two ways of obtaining an answer from the data in the contract. They differ in whether a new logical fact must be formed. In the static case, all data constituting the answer are present in the contract from the outset. In the dynamic case, the answer requires a new logical fact to be formed from the initial data. The relevant operations include counting actions, generalizing over them, and drawing conclusions subject to the conditions in the question. Both cases are considered below. Passages of the proofs specific to the dynamic case are marked (*).

K3. A full answer supplies all the information requested, subject to the conditions of the question. Objects specified in the question retain their meaning in the answer, and their designations can be checked unambiguously against the contract.

K4. If the answer is abbreviated, its full content must be uniquely recoverable from the question and contract. For example, in response to “Where is David?”, “In the park” means “David is in the park.”

K5. The full answer is justified by one or more selected facts of the contract. When sufficiency is checked, other information in the contract is not added automatically. Any intermediate fact whose content is needed to justify the answer must be explicitly accounted for.

K6. The answer is checked in one fixed state of the contract, either the initial state or one extended by derived facts. A new fact enters the contract on the strength of a verified inference from facts already available. Its premises and the rule applied are retained for verification. The state of the contract remains fixed throughout the preserving transformation $S \to S'$.

K7. W is the set of facts in the contract found by searching from the objects in the question. It is sufficient for the question if the facts in W and the verified connections between them establish all the requested relations with the specified participants, roles, and conditions. For a given answer we select a nonempty set $S \subseteq W$ whose facts jointly justify those relations. S includes the parent and intermediate facts, inclusions, and targets of preserving references used in the justification. The selected justification must contain a verifiable designation of an object common to the question and the full answer. Evidence for resolving designations is handled under S6.

## Rules for Constructing a Link

P1. The underlying fact of a link is a logical fact in the selected fixed state of the contract. When a nested subfact is selected, the subfact itself is the underlying fact of the link. Every fact used has an admissible finite initial textual representation. A full textual representation preserves the content of the selected fact and the designations of its objects. Shortening is permitted under P3.

P2. Local content-preserving expansion. In an admissible finite representation of a parent, a reference to a particular established fact may be replaced, at the same position in the meaning of the parent text, by an admissible full finite representation of that fact. An abbreviated occurrence of an established subfact may be expanded in the same way. The operation preserves the full content of the containing construction, its participants, the identity of events, relations, roles, conditions, and their scope, as well as the targets of the remaining references and any required designations already presented. The resulting text remains an admissible representation of the same parent, so the rule continues to apply after earlier expansions. Grammatical adjustment and the use of several sentences are permitted. New events may not be introduced, and relations may not be changed.

P3. Textual surgery shortens a text while keeping its underlying fact fixed. It preserves the required part of the content already asserted, together with its meaning, status, roles, and necessary conditions. The shortened text follows from the original text without additional premises. Shortening may neither turn doubt or supposition into a statement that an event occurred nor remove conditions that bind the assertions together. Selecting another subfact is a separate operation. A shortening is admissible if, with the other links fixed, it preserves the link’s contribution under P5, the justification of the full answer, the necessary connections, and verifiable object designations, including the object common to question and answer.

P4. A lexical key connects verifiable designations of the same object in adjacent elements of the chain. These designations may differ in wording. The object designated may occur in the underlying fact itself or in its subfacts. The designations selected for transitions are retained in the link text. The contract uniquely determines the object designated by each key. A reference must be traceable to the object in finitely many steps, each confirmed by the contract.

P5. The content of a link contributes to the joint justification of the full answer. It may supply a requested piece of information, establish objects, conditions, or relations needed for the answer, or connect necessary facts through relations and references confirmed by the contract. This contribution is assessed within the selected chain, preserving the participants and their roles. A link need not state the final answer or a self-contained part of the information requested. Matching designations alone do not constitute the required contribution.

P6. A link text has finite length. It is minimally sufficient in the selected chain if, with all other links fixed, it admits no further shortening under P3. Length is measured in characters in a fixed textual representation. No common upper bound on text length is imposed across links.

## Rules for Constructing a Chain

S1. A chain from the question to the full answer is a linear sequence of a finite number of links, at least one.

S2. The same selected fact cannot serve as a link twice in one chain. Different facts may serve as different links, including distinct subfacts of the same parent, or a parent and its subfact. Repetition is determined by the particular fact selected for a link, independently of a shared source or a partial overlap of content.

S3. Different links may share objects, and their content may overlap. The same key may be used in consecutive transitions.

S4. Each pair of adjacent links must meet two conditions: an admissible structural connection between their underlying facts in the original S, and verifiable designations of the same object in their texts. Structural connection is established through preserving inclusions, references, or existing unifying parent constructions, with evidence for the relevant relations. A shared lexical key alone does not establish structural connection. At the boundaries of the chain, the question connects to the first link and the last link to the full answer through verifiable lexical keys. Keys are checked under P4.

S5. The contents of the selected links must jointly justify the full answer under K5. Chain connectivity and sufficiency for the answer are checked separately.

S6. A fact used solely to resolve a reference remains available as verifiable evidence of the correspondence between designations under T9.2. It need not become a separate link. If its content is also needed to justify the answer, the fact is included in the selected justification, and its content is retained in the link texts. A fact may serve both purposes. Its use in resolving a reference does not remove the requirement to retain its contribution to the justification.

## Lemma 1. A Common Lexical Key for the Question and the Full Answer

*"If an answer is full and sufficient for the question asked, its full content is connected to the question by at least one lexical key."*

$$\mathrm{Full}(T_0, Q, A) \Rightarrow \exists\, K_0 : Q \leftrightarrow(K_0 \mid T_0)\ \hat{A}.$$

The claim also holds when facts sufficient for the answer have been derived by a justified inference and added to the contract, provided the specified objects are preserved and their designations are unambiguous in the selected extended state. The lemma applies to a single fixed state after the derived facts have been added.

*Notation*

- C is the source text, Q the question, and A the answer.
- $\hat{A}$ is the full content of the answer. For example, the answer "In the park" to "Where is David?" expands to "David is in the park."
- $\mathrm{Full}(T_0, Q, A)$ means that A provides all the information requested, preserving the participants and conditions of the question. Its full content is recoverable from the question and contract and is justified by facts of the initial contract $T_0$ (K3–K5).
- Resolve(e) = o means that the contract $T_0$ establishes that the expression e designates the object o. Here Resolve = $\mathrm{Resolve}_{T_0}$.
- $Q \leftrightarrow(K_0 \mid T_0)\ \hat{A}$ means that the question and the full answer contain designations of the same object, as confirmed by $T_0$.

• $K_0$ is the designation of that object in the question together with the evidence connecting it to the designation in the full content of the answer Â.

• $Obj_T(V)$, or Obj(V) for fixed T, is the set of objects with verifiable designations in text V. $Resolve_T(e) = o$ means that, under T, the designation e refers to object o.

$T_0$ is the initial contract, including the main context and its interpretive frame (T1–T6). $F_0$ is the set of its initial facts. $D_t$ is the finite set of derived facts added, with $D_0 = \varnothing$. $T_t$ is the state of the contract after t addition steps. The initial information, the contract frame, and the meanings of the designations are preserved.

$Full^*(T_t, Q, A)$ means that the initial and derived facts in $T_t$ justify the full and sufficient answer A, and that the inferences used can be traced back to the initial facts in $T_0$. Recovery of Â and verification of designations take place in the same state. Here $Resolve_t = Resolve_{T_t}$. The expression $Resolve_t(e) = o$ means that e resolves unambiguously in $T_t$ on grounds verifiable in $T_0$. $Q \leftrightarrow (K_0 \mid T_t)\ \hat{A}$ denotes a connection between the question and the full answer through such a shared object.

*Proof.*

Consider a question Q and an answer A that supplies all the information requested, preserves the participants and conditions of the question, and is justified by facts of the initial contract $T_0$ (K3, K5). Write Â for the full content of the answer. If the answer is abbreviated, that content is uniquely recoverable from the question and the contract (K4).

(*) Consider a full and sufficient answer in state $T_t$. The facts justifying it may be initial or derived. Each new fact d is added by a verified inference from a finite set Γ of facts already available:

$$\Gamma \subseteq F_0 \cup D_j, \quad |\Gamma| < \infty, \quad \Gamma \vdash d,$$
$$D_{j+1} = D_j \cup \{d\}, \quad 0 \leq j < t.$$

(*) At the first step the premises belong to $F_0$. Suppose all previously added facts have justifications based on $F_0$. Substitution of those justifications into the inference of d then gives a justification of d from $F_0$. By induction this holds for every added fact. Thus an answer sufficiently justified in $T_t$ remains grounded in the initial information:

$$\forall\, d \in D_t\colon F_0 \vdash d,$$
$$S \subseteq F_0 \cup D_t, \quad S \vdash \hat{A} \quad \Rightarrow \quad F_0 \vdash \hat{A}.$$

(*) Here S is a finite set containing the facts selected to justify the answer. The premises and rules used in every inference are retained. Once $T_t$ has been fixed, the proof proceeds without adding further facts.

The question refers to at least one object (K1.1). Write o for one such object. The full answer preserves the identity of that object, and the correspondence between its designations is verifiable in the contract (K3). For an abbreviated answer, we use the recovered full content Â (K4). The designations in the question and the full answer can therefore differ in wording while referring to the same object o.

Let Obj(X) be the set of objects designated in X. For the selected object we obtain

$$o \in Obj(Q) \cap Obj(\hat{A}).$$

Let $e_Q$ designate the selected object in the question and $e_{\hat{A}}$ designate it in the full answer.
The initial contract $T_0$ unambiguously establishes that both designations refer to the selected object (K3). This gives

$$Resolve(e_Q) = o = Resolve(e_{\hat{A}}).$$

(*) In the fixed state $T_t$ the objects of the question are preserved, and an abbreviated answer is restored to its full content (K3–K4, K6). The designations $e_Q$ and $e_{\hat{A}}$ are checked over all interpretations admissible in this state. If the check uses a derived fact, its justification and the link

between the designation and the initial object are traced back to $T_0$. The fact is recorded as the result of an inference. It does not count as a quotation from C. We obtain

$$\mathrm{Resolve}_t(e_Q) = o = \mathrm{Resolve}_t(e_{\hat{A}}).$$

The expression $e_Q$, together with the connection to $e_{\hat{A}}$ confirmed in $T_0$, forms a common key $K_0$ between the question and the full content of the answer:

$$\mathrm{Full}(T_0, Q, A) \Rightarrow \exists\, K_0\text{: } Q \leftrightarrow(K_0 \mid T_0)\ \hat{A}.$$

This proves the lemma under the stated conditions. □

(*) The established correspondence of the designations to one object also supplies a common key in the extended state:

$$\mathrm{Full}^*(T_t, Q, A) \Rightarrow \exists\, K_0\text{: } Q \leftrightarrow(K_0 \mid T_t)\ \hat{A}.$$

(*) This proves the dynamic extension under the stated conditions. □

## Theorem. Existence of a Chain of Facts from Question to Answer

*“If the initial context contains a full and sufficient answer to a question, it contains at least one linear chain of logical facts connected by lexically expressed keys that links the question to the answer and justifies the answer.”*

An answer is full if its content covers all the information and relations requested by the question. A justification is sufficient if that content follows from the selected facts of the contract. Full, abbreviated, and synonymous constructions correspond to one logical fact when, under the question and contract, they express the same assertion with the same participants, roles, and conditions (T8.4, K3–K5).

In the hypothesis of the theorem, “the context contains a full and sufficient answer” means that the contract contains a logical fact r that expresses the requested relation and justifies the full answer Â. If the content of the facts nested in r is sufficient for the answer, the answer may be presented through those facts. Presenting them as separate links requires an order and link texts that preserve the justification of the full answer and for which all transitions are verified (§§4.3–4.4). Constructing a chain requires that the selected expansion of r provide a verifiable designation of an object common to the question and the full answer (K7, P4). The conclusion of the theorem is written as follows:

$$\exists\, L=(Z_1,\dots,Z_n)\in\mathcal{G}(T,Q,A)\ \exists\, o\text{: } o\in \mathrm{Obj}(Q)\cap \mathrm{Obj}(\hat{A})\cap \textstyle\bigcup_{i=1}^{n} \mathrm{Obj}(V_i).$$

(*) The dynamic extension concerns a fixed state of the contract after a finite sequence of justified inferences and registrations of facts. Both graph construction and verification of the boundary key are relative to this state.

*Notation*

- C, Q, A, Â, $K_0$, Obj, and Resolve have the same meanings as in the lemma on the common lexical key for the question and the full answer.
- T is the selected fixed state of the contract, with its contents specified under T1–T6. In the static case $T=T_0$. In the dynamic extension $T=T_t$. $\mathrm{Obj}_T$ and $\mathrm{Resolve}_T$ refer to the selected state. The subscript may be omitted when T is fixed.
- $\mathrm{ResolveFact}_T(f) = F$ means that under T the name f uniquely designates the logical fact F. This notation is used in the section on the fact registry. $\mathrm{Resolve}_T(e) = o$ applies to object designations.
- L is a chain from the question to the full answer: $Q \leftrightarrow Z_1 \leftrightarrow \dots \leftrightarrow Z_n \leftrightarrow \hat{A}$. The number of links n is finite and at least one.
- $Z_i = (P_i, V_i)$ is a link with underlying fact $P_i$ and admissible textual representation $V_i$. The designations used as keys occur in $V_i$.
- Each transition $\leftrightarrow_x$ connects verified designations of a single object x in adjacent elements of the chain. The incoming and outgoing transitions may use different objects.

- Facts(T) is the set of existing logical facts in the selected contract. Equivalent lexical expressions under T8.4 count as one fact. p(V) is the content of an admissible textual representation V. For a family X of representations, p(X) denotes the set of their contents. When premises are combined by union, an individual assertion p(V) stands for the singleton set {p(V)}.
- $Ref_T(F,G)$ denotes a confirmed reference from F to fact G. $Ref^+_T(F,G)$ denotes a reference admitting full content-preserving expansion under P2. Resolving the target alone does not suffice for such an expansion. A reference edge in the selected state T corresponds to a reference $Ref^+_T(F,G)$.
- $P \succcurlyeq_c F$ means that P preserves the full asserted content of F in the sense of T9.4. The subscript c distinguishes content preservation from graph reachability.
- $P \succcurlyeq_G F$ means that there is a directed preserving path from P to F in the graph G under consideration, allowing a path of length zero. $P \succ_G F$ requires positive length. The subscript is S for the justification graph and K for the contract graph.
- $Rel_S(F,H)$ means that the graph of the selected S contains a path without repeated vertices from F to H that first runs against the direction of the preserving edges and then follows their direction. Either segment may be empty. The edges have witnesses in the graph. At an internal turn the parent expresses the relation between the participants concerned. For F = H a path of length zero is admissible (T8.3, T9.4, T10.1, S4).
- Key(V,W) means that the texts V and W contain verifiable designations of the same object under P4.
- $\Gamma \vdash B$ denotes logical entailment from selected premises: $\Gamma \vdash B \Leftrightarrow \forall I: ((\forall F \in \Gamma: I \models F) \Rightarrow I \models B)$. Here I is an admissible interpretation of the assertions, with the meanings of the designations and the rules of logic fixed. The condition is checked over all such interpretations satisfying the selected premises Γ, including those that do not satisfy the entire contract. A missing premise therefore cannot be supplied implicitly from the rest of the text. $F \vdash B$ denotes $\{F\} \vdash B$. The notation $B \Leftrightarrow D$ denotes entailment in both directions. For a set S of logical facts, $S \vdash \hat{A}$ takes their contents as premises. For a family Y, $\Gamma \vdash Y$ means that every member of Y is entailed. A chain justifies the answer if $\{p(V_1), \ldots, p(V_n)\} \vdash \hat{A}$.
- $\mathcal{J}(T, Q, A)$ is the family of chains that justify the full answer and satisfy the conditions on links and transitions. Minimality of the chain as a whole is not required.
- $F \preccurlyeq P$ means that F is preserved in P as an asserted subfact, directly or through nested subfacts. Equality F = P is allowed. P entails F.

# 1. The Question and the Content of the Answer

## 1.1. The Question and the Root Fact of the Answer

### Proposition 1.1. How the Question Constrains the Root Fact of an Answer

*“The question specifies the requirements on the content of a possible root fact for the answer. It indicates which relation is to be established and under what conditions. If a logical fact in the selected state of the contract expresses this relation and justifies the full answer, it can serve as the root fact of the selected analysis.”*

The question may name participants in the relation or specify the role of a participant whose identity remains to be established (K1.1). The answer must preserve the meaning of these indications and the conditions of the question (K3). If the participants are themselves facts, their relation is expressed by a parent fact. Its nested facts form the first level of the expansion of the root (T10.1).

In the graph of a composite answer, the parent fact preserves the participants and expresses the relation between them. If the joint content of its subfacts establishes all the information and relations requested, the answer can be presented through those subfacts. The parent remains in the justification graph. If only the parent expresses the requested relation, the expansion used to present the answer must preserve that relation (T10.1, K3, K5).

For example, in the question “Who had access to the operating system in the inner room of the data center?”, denote this access by o. The access in question is known, and the task is to identify its holders. The contract supplies the participants who occupy that role. Several holders may satisfy the same condition in the question.

If the question admits a division into independent questions that preserves the requested relations, a root fact is considered separately for each of them (K1.2).

## 2. Graph Structures of the Contract, the Question, and the Answer

## 2.1. The Contract Graph

The contract comprises the information and conditions of interpretation on which the justification of an answer rests. It includes the main context and the frame. The main context contains messages and assertions. The frame specifies the language, participant designations, headings, source structure, dates, and other necessary information (T1–T2).

We consider the content of the context together with these conditions. For example, a heading may establish the period during which the actions in a list took place, and dialogue markup may identify the speaker of an utterance. When identifying facts, we preserve the relations expressed in this way and their association with the events concerned. The grounds for interpretation remain linked to the messages and assertions to which they apply (T3–T6).

*Logical facts and their representations*

Each logical fact represents an accepted assertion with particular participants, relations, and conditions. It corresponds to one vertex in the graph. Full, abbreviated, and synonymous constructions designating this fact form its family of lexical representations (T8–T8.4).

For example, suppose the contract establishes that “David went to the library” designates the full fact “David went to the library late.” Both constructions then correspond to one vertex. The temporal qualification remains part of the full content of the designated fact. The contract establishes the correspondence between the abbreviated and full expressions (T9.2.2, T10.2).

Write K for the set of logical facts of the contract and $L_T(F)$ for the family of lexical representations of fact F. The model assumes that there are finitely many logical facts (T12).

$K = \mathrm{Facts}(T), \quad |K| < \infty,$

$L_T(F)$ is the family of lexical representations of fact F.

The construction “Emma made dinner and then set the table” contains two facts about Emma’s actions and establishes their order. The parent and each of the two subfacts have their own vertices in the graph (T10–T10.2).

Object designations occur in the content of facts. Verifiable designations of the same object in different texts serve as lexical keys (T8.2, P4).

*Preserving connections*

A connection between logical facts can be represented in three ways: by nesting one fact within another, by a preserving reference, or through a common parent that expresses their relation (T9.1, T8.3, T10.1, K7).

- Inclusion. One fact includes the full asserted content of another. A directed edge runs from the containing fact to the nested fact. The assertion, participants, their roles, and the conditions are preserved (T9.1, T9.4).

  $X \to Y.$

- Preserving reference. The referring fact preserves the content of its target. Establishing this connection requires identifying the target and checking that the reference can be expanded into a full representation of the target while preserving the content of the entire construction, its conditions, and the status of the nested assertion (T8.3, T8.5, P2).

$X \dashrightarrow Y$.

- Common parent. The parent preserves both subfacts and expresses their relation. Two directed edges from the parent to the participants in the relation represent the connection between the subfacts (T10–T10.1, K7).

$X \leftarrow R \rightarrow Y$.

The arrow ⇢ distinguishes a preserving reference. In paths it counts as a preserving edge → with the same direction. Each inclusion or reference is supported by content that confirms it. For a parent construction we also record the expressed relation and the roles of its participants. Together with the addresses of the occurrences, these data constitute the witnesses for the connections. For each edge used in expansion, the witness identifies the address of the subfact occurrence in a full representation of the parent.

We use the following notation for the preserving graph of the contract:

- $G_K$ is the preserving graph of the contract.
- K is its set of logical facts.
- $E_K$ is its set of strict directed edges between distinct facts.
- $\lambda_K$ records the witnesses for inclusions, preserving references, and parent relations, including participants, roles, and addresses.

The set of facts, directed edges, and witnesses for their connections determine the graph:

$$G_K = (K, E_K, \lambda_K),$$
$$E_K \subseteq \{(X,Y) \in K \times K : X \neq Y\}.$$

*Geometry of the connections*

The contract is represented by a geometric structure of connected logical facts. Parent constructions combine nested facts into progressively larger assertions. These constructions may converge on one root fact or give rise to several roots.

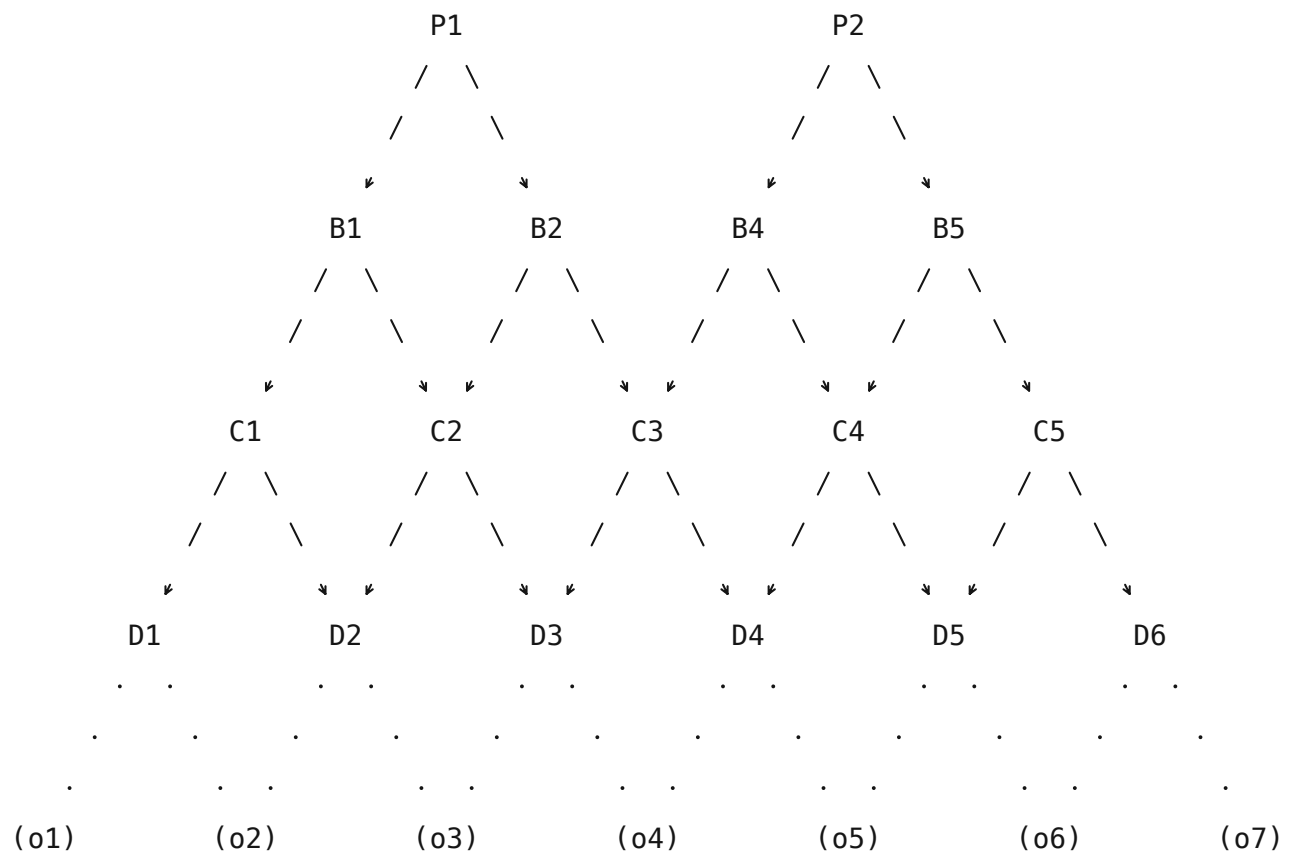

**Figure 2.1.** Geometric structure of a contract

*Notation*

P1 and P2 are the root facts of the graph shown.

B1, B2, B4, and B5 are parent facts.

C1–C5 are nested facts. C2, C3, and C4 each belong to two parents.

D1–D6 are nested facts at the next level.

The arrows represent preserving inclusions of facts.

Dots below the D facts indicate that their content includes designations of objects.

(o1)–(o7) are objects.

Each composite fact at level B or C is a parent that preserves its nested subfacts and expresses relations between them. The root facts at level P have no further preserving parent in the contract graph under consideration. In this structural sense, the level-P facts are recorded as standalone assertions in the contract.

The lowest level of the diagram consists of objects o, the atomic units of the chosen analysis. Facts express the properties and relations of these objects established in the contract. Successive logical levels combine facts and express relations between them, up to the root facts.

(*) If new facts have been derived for the answer, we first add them to the contract with their verified grounds of inference and fix the extended state. We then represent its facts and preserving connections in the same way. For each new fact we separately retain its provenance: the premises and the inference rule applied. Preserving edges to premises are included when the result is confirmed to preserve their full content.

## 2.2. Representing the Question in the Contract Graph

The question indicates which participants in the requested relation are already known and what remains to be established about them. If it names an established fact, the designation corresponds to a vertex in the contract graph. If it indicates an object, its designation is matched to designations of the same object in the content of facts. Both direct names and references with a confirmed target are taken into account (K1.1, T9.2).

Suppose the question "Why did David leave?" designates an established fact of David's departure. One vertex in the graph is thereby identified. The word "why" specifies the direction of the relation: the departure occupies the role of effect, and its cause is sought. "What happened after David left?" points to the same vertex but assigns the departure the role of earlier event. Thus questions can indicate the same vertex while imposing different requirements on the answer.

In the access question in §1.1, only the object o is known. The requirement on participant x can be written as an open expression:

q(x) = "x had the access designated by o."

Here x marks the place of the access holder. For every value of x included in the answer, the corresponding assertion must be justified in the contract. The open expression itself does not yet specify a vertex with a particular holder. The information supplied by the question can therefore be limited to an object designation and the role in which a participant must be identified.

## 2.3. The Answer Graph and Its Set of Facts S

Suppose the relation requested by the question is expressed by a fact r in the selected state of the contract. To analyze this answer, first take r itself. If r is composite, identify the facts between which it establishes a relation. Each can be expanded in the same way. For an inclusion, proceed to the subfact. For a preserving reference, proceed to its target. In both cases the containing fact preserves what the nested or referenced fact asserts, together with that fact's role and the conditions under which its assertion holds (T8.3, T9.4, T10.1).

Let S denote the set of logical facts identified in this analysis. Each such fact corresponds to a vertex. An edge points from the containing fact to the fact it preserves. A subfact participating in two parent relations has one vertex with two incoming edges. For each occurrence, a witness records which parent expressed the relation and what role the subfact played (T8.4, T10.1).

Let $E_S$ be the edges used in the expansion and $\lambda_S$ their witnesses from the contract. The answer graph has the form

$$\Gamma_S = (S, E_S, \lambda_S),$$
$$r \in S \subseteq K, \quad E_S \subseteq E_K \cap (S \times S).$$

**Proposition 2.3. The Root Fact of the Selected Analysis**

*"The expansion graph of the fact r is finite. Every fact in the selected analysis is reachable from r, and r is its unique root."*

*Proof.* Initially the analysis contains only r. Expanding a fact P already present adds a participant F along a verified preserving edge. Appending this edge to the path from r to P gives a path from r to F. Thus all facts identified at every step are reachable from r. The number of distinct vertices is bounded by the number of facts in the contract (T12).

For every $F \in S$ there is a path $r \to \ldots \to F$ in $\Gamma_S$.

Each fact other than r has an incoming edge, the last edge on such a path. An edge entering r itself would create a cycle of strict preserving edges. Distinct facts on the cycle would then fully preserve one another. Mutual full preservation implies that their asserted content, participants, roles, and conditions coincide. They would therefore be one logical fact. Since a strict edge connects distinct facts, such a cycle is impossible (T8.4, T9.4). The only root is r. □

In Figure 3.2, whose root is B1, E1 belongs to both C1 and C2. E1 occurs only once in the answer graph, and both parent relations are retained. B1 is a root relative to the selected analysis. In the contract graph it may itself belong to a larger fact.

*Limits of expansion from a root fact*

Let $D_K(r)$ denote the set of contract facts accessible under full expansion of r. Each step follows a preserving edge from parent to subfact:

$$D_K(r) = \{F \in K : r \succcurlyeq_K F\}.$$

If r preserves Y, any preserving path from Y can be extended to a path from r by prefixing the path from r to Y. Full content is preserved along the resulting path (T9.4). It follows that

$$r \succcurlyeq_K Y \Rightarrow D_K(Y) \subseteq D_K(r).$$

In the graph of the selected analysis, these paths are restricted to edges in $E_S$. By Proposition 2.3 they reach all the facts in S from r. Writing $D_S(r)$ for the set accessible in this graph, we have

$$D_S(r) = S \subseteq D_K(r) \subseteq K.$$

Equality $D_K(r) = K$ requires that every fact of the contract be reachable from r along a directed path. In particular, suppose K is covered by the sets $D_K(r_i)$, and each $r_i$ is reachable from r. Then $D_K(r_i) \subseteq D_K(r)$ for each i, which gives $D_K(r) = K$.

(*) A derived fact is expanded after it has been registered in the selected state of the contract. A premise of an inference enters the expansion graph via a preserving edge only when the resulting fact preserves its full content (K6, T9.4).

## 3. Searching for Logical Facts and Growing the Tree of Relations

### 3.1. Searching for Logical Facts in the Contract

The unit of search is a logical fact. All lexical constructions expressing the same fact in the contract correspond to one vertex (T8.4). Any established representation of the fact can therefore be used to find its occurrences in other assertions, through nesting or a preserving reference. The theoretical model takes the correspondence between lexical representations and logical facts as established.

Search ranges over the entire fixed contract T. A query for an object o returns the facts in which a designation of o is confirmed. For a fact F already found, the query returns every fact P that directly includes F or preserves its content through a reference. Expanding P in the opposite direction gives access to its immediate subfacts and the targets of its preserving references.

The operations use object designations and edges of the contract graph together with their witnesses. The set S, its root fact, and labels indicating which facts belong to the answer graph are not inputs to the search. Membership of a search result in the answer graph must therefore be justified separately.

## 3.2. From an Object in the Question to the Initial Facts of the Contract

In the fixed contract T, the question Q specifies what information is sought and what conditions the answer must satisfy. At least one object of the question is designated directly or through a confirmed reference (K1.1). Write $\mathrm{Obj}_T(Q)$ for the set of such objects. Then

$$\mathrm{Obj}_T(Q) \neq \varnothing.$$

For a logical fact F, we determine $\mathrm{Obj}_T(F)$ from the expressions in $L_T(F)$ recorded in the contract. An object o belongs to $\mathrm{Obj}_T(F)$ if at least one of those expressions names it directly or designates it through a confirmed reference (T8.4, T9.2). We use this condition to select the initial facts before expanding any nested facts or preserving references:

$$I_T(o) = \{F \in \mathrm{Facts}(T) : o \in \mathrm{Obj}_T(F)\}.$$

If several objects are designated in the question, the search starts from all facts containing at least one of them. Write $I_T(Q)$ for the resulting set:

$$I_T(Q) = \bigcup_{o \in \mathrm{Obj}_T(Q)} I_T(o).$$

The question refers to at least one object, and for each such object the contract contains a fact that designates it (K1.1, T11). Hence $I_T(Q)$ is nonempty. Its finiteness follows from the finiteness of the set of contract facts (T12).

In the access example, a designation of o is recorded in a presentation of E8, so E8 belongs to $I_T(o)$. If o is also designated in presentations of C11 or C10, the corresponding parent facts are found immediately. If the designation occurs only in E8, both parents are discovered by ascending from E8 along preserving edges. Whether the facts found belong to the justification of the answer is checked against the relations they express.

(*) In the dynamic case the search takes place in the selected state $T_t$ after the derived facts have been registered. $I_T(Q)$ is defined using the facts and designations of that state.

## 3.3. Ascending to Candidate Root Facts

For each initial fact F, find every contract fact P that includes F or contains a reference to F admitting full expansion. The arrow P → F points from the preserving fact to the preserved fact. Search ascends along this edge from F to P. Apply the same operation to each P found. If several branches lead to the same fact, its parents need only be checked once.

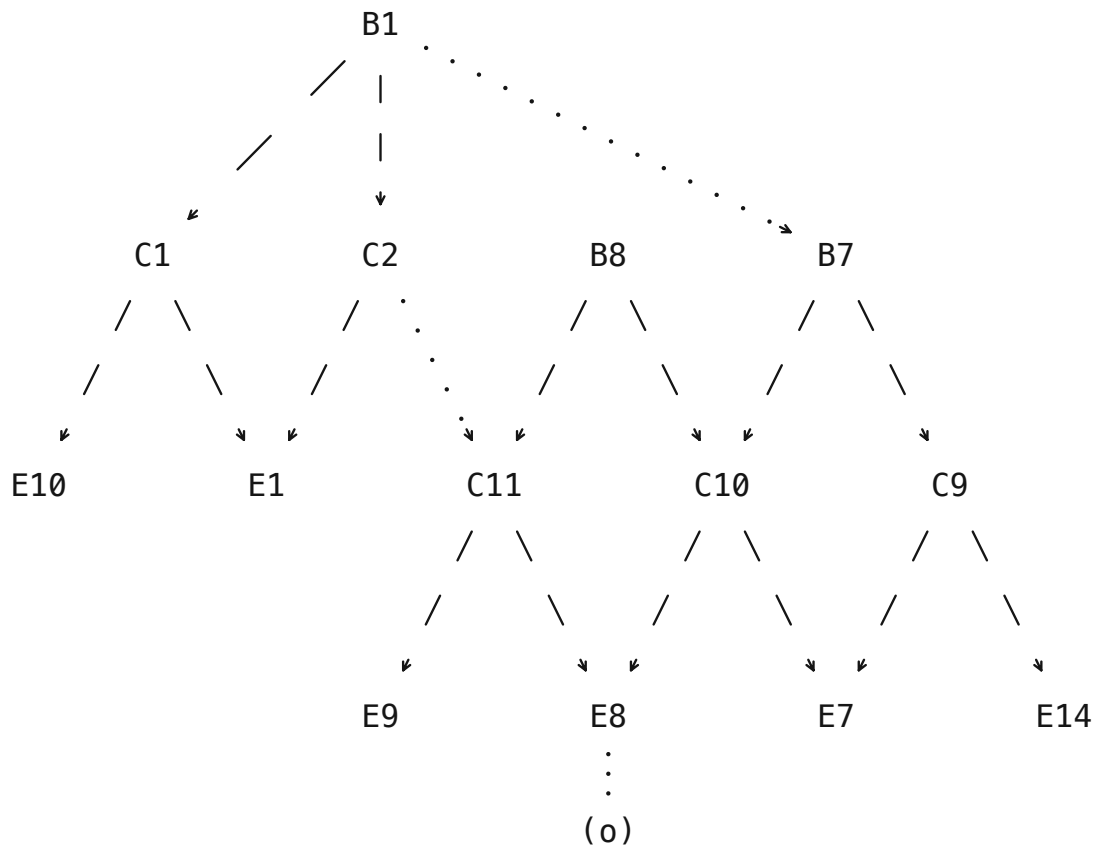

**Figure 3.1.** Ascent from E8 in the contract graph (fragment)

Arrows point from parent to subfact. Strokes indicate inclusions, and dots along an arrow indicate a preserving reference. Below E8, (o) denotes the object "access."

Search ascends from E8 to C11 and C10. On the left, C2 connects E1 and C11. On the right, B7 connects C10 and C9. Both branches lead to B1. In the same fragment, B8 directly preserves C11 and C10. There is no direct edge between B1 and B8.

Let $U_n$ be the set of facts found in at most n steps of ascent. The initial set and one expansion step are given by

$$U_0 = I_T(Q),$$
$$U_{n+1} = U_n \cup \{P \in K : \text{there exists } F \in U_n \text{ such that } (P,F) \in E_K\}.$$

**Lemma 3.3.1. Completeness of Ascent along Preserving Edges**

*"In a finite contract graph, ascent finds precisely the facts from which a preserving path leads to at least one initial fact. An initial fact is included by a path of length zero."*

*Proof.* After zero steps only the initial facts have been found. Suppose that after n steps we have found precisely the starting vertices of paths of length at most n ending in $I_T(Q)$. Each newly found P preserves a fact F already found. Prefixing the edge $P \to F$ to the path from F to an initial fact gives a path from P of length at most n + 1.

Conversely, the first edge on any path of length n + 1 leads to a fact F from which n edges remain to an initial fact. By the induction hypothesis F has already been found. The next expansion finds P. The claim follows for every n. As long as the discovered set grows, it gains at least one element of the finite set K. Once a step adds no facts, further applications of the rule leave the set unchanged. □

(*) For a fixed extended state $T_t$, ascent follows the preserving edges of that state. An inference premise is connected to the result by such an edge only if there is a confirmed inclusion or preserving reference.

## 3.4. Growing the Tree of Relations

We call the successive discovery of contract facts along preserving edges growing the tree of relations. The process can start from a fact found through an object of the question, regardless of where that fact lies in the graph. For each discovered fact, find both its subfacts and the parents that preserve it. The result is a graph. If a subfact belongs to two parents, it has one vertex with two incoming edges. Each edge retains its own witness.

Write f for the initial fact and $W_n$ for the set of facts discovered in at most n steps. One step is given by

$$W_0 = \{f\},$$
$$W_{n+1} = W_n \cup \{Y \in K : \text{there exists } X \in W_n \text{ such that } (X,Y) \in E_K \text{ or } (Y,X) \in E_K\}.$$

Edges are traversed in both directions. In the resulting graph each edge retains its original direction and its witness. For example, starting from a shared subfact reveals both parents, but that step alone does not establish a relation between the parents.

**Proposition 3.4. Completeness of Growing the Tree of Relations**

*"In a finite contract, growing the tree from f discovers all and only the facts reachable from f along preserving edges when traversal may change direction."*

*Proof.* Initially only f has been discovered. Suppose that after n steps all facts at distance at most n have been found. Each new fact is connected by a preserving edge to a fact already discovered and so is reachable in n + 1 steps. Conversely, on any route of length n + 1, the penultimate vertex is reachable in n steps. Checking its incident edges discovers the final edge. Induction gives the claim for routes of any finite length. The construction ends when all connections of the discovered vertices have been checked and no new vertex is found. Since K is finite, this stopping condition is reached (T12). □

Let W denote the set of facts discovered by the time the process stops. The graph on W contains every preserving edge of the contract with both endpoints in W. For multiple initial facts, take $W_0 = I_T(Q)$. If one initial fact is reachable from another by traversing edges in either direction, both discover the same component. If initial facts lie in different components, W contains the vertices of each of those components. As soon as the initial set contains a fact from a connected analysis S, growing the tree covers all of S. Other branches of the contract may also be discovered.

Within the fragment in Figure 3.1, traversal in both directions from E8 reaches all 14 facts shown. Through C11 and C10 the search discovers B8, and along the branches through C2 and B7 it reaches B1. In Figure 3.2, expansion follows the arrows from B1 to its subfacts and covers 13 facts, including B1 itself. B8 lies outside this expansion: it contains C11 and C10, but the graph shown has no directed path from B1 to B8.

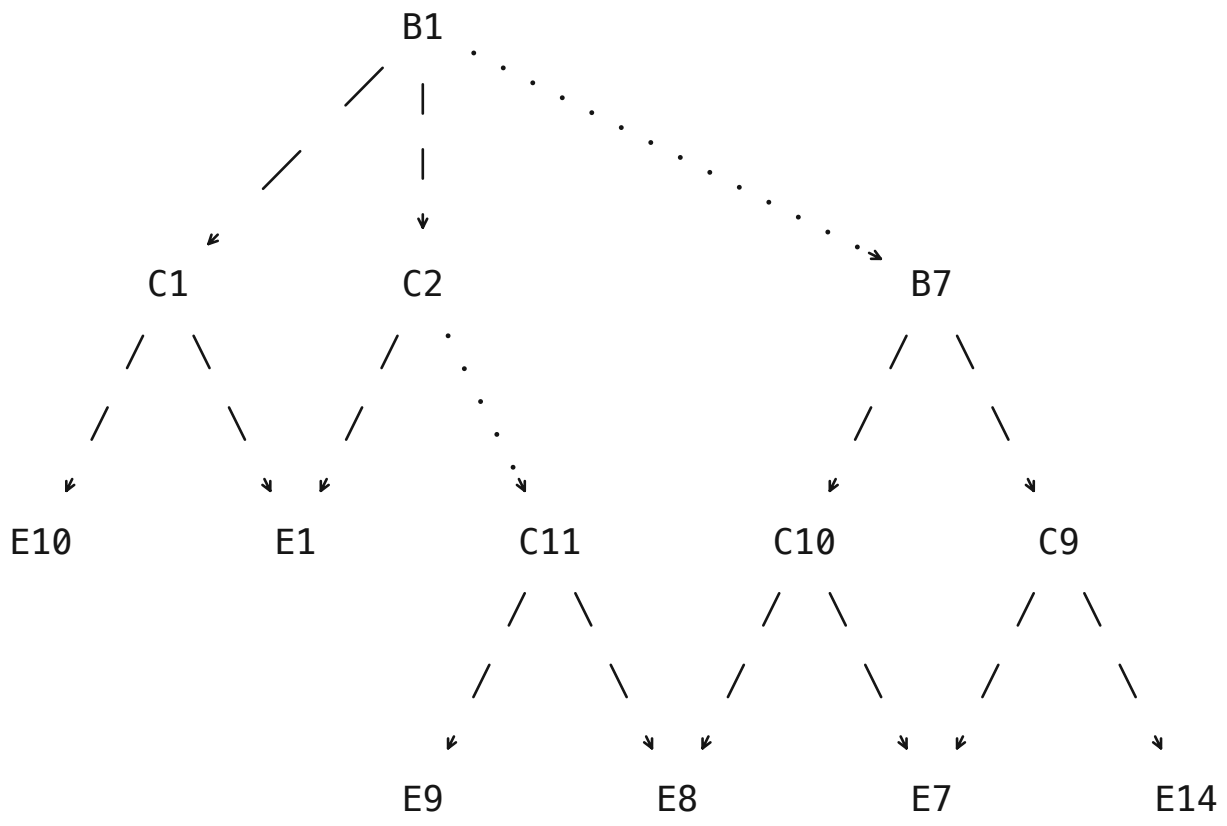

**Figure 3.2.** Expansion graph of the root fact B1

The notation is the same as in Figure 3.1. Expansion follows the arrows from B1. Each shared subfact (E1, E8, E7) appears as one vertex with two incoming edges.

In the graph of discovered facts, two root facts may share a subfact. In the following fragment, growing the tree from B discovers both P and Q, whereas expanding P along its arrows discovers A and B.

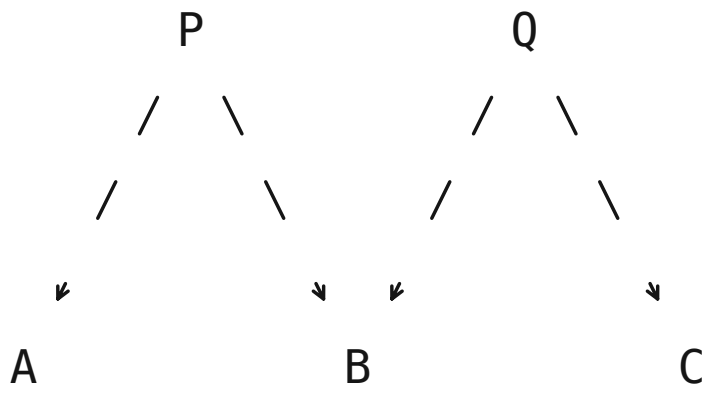

**Figure 3.3.** A shared subfact B of two root facts P and Q

There is no admissible structural connection within this fragment between the two distinct roots. Such a connection would have the form P ← … ← r → … → Q, including the case where r is one of the endpoints. At least one of the two paths would have positive length, and its last edge would enter a root, contrary to the definition of a root. Thus discovering P and Q through B during growth does not establish the relation between the whole facts P and Q required for adjacent links (S4). If another parent in the contract preserves both facts and expresses their relation, P and Q are no longer both roots in the larger graph.

(*) In the extended state, growth follows the preserving edges of registered facts. The inference log retains the premises and the rule applied. Moving from a result to a premise along a graph edge requires a confirmed inclusion or preserving reference by which the result preserves the full content of that premise (K6, T9.4).

# 4. Constructing Chains in the Answer Graph

The question determines which relations must be established. W is sufficient for the question if the facts found and the connections between them establish all those relations with the specified participants, roles, and conditions. For each required relation we identify the facts in W and the verified connections that express it. The number of branches alone does not affect the sufficiency of W (K1.1, K3).

In the graph on W, select a justification S for one answer. Let r be the root fact of the selected analysis. First take a chain with a single link whose underlying fact is r. It preserves the participating facts and the relations expressed between them. Once an admissible text has been chosen and the boundary key verified, this link connects the question to the full answer (§§4.1–4.2). We then consider more detailed presentations of the same answer by replacing one parent fact with several nested facts and checking their order (§§4.3–4.4).

For B1 in Figure 3.2, the successive expansions can be written as sequences of underlying facts:

```
       [B1]
        |
        |  Expand B1
        ↓
   [C1, C2, B7]
        |
        |  Expand B7
        ↓
 [C1, C2, C10, C9]
```

**Figure 4.1.** Expansion of links B1 and B7

Here the arrows denote link replacement. Under the conditions of §4.2, the first line gives the initial single-link presentation. The second and third lines are candidates to be checked. Each is checked independently. Acceptance of the third does not depend on prior acceptance of the second.

*Selecting a root fact in the graph of discovered facts*

W may contain several parent facts. For each one we check whether it expresses the requested relation with the required participants and conditions. Consider the question "Who had access to the system?" Suppose the contract establishes the parent fact r: "Only Emma, David, and Kyle had access to the system." Three subfacts can be distinguished in it.

$F_1$ — Emma had access to the system.
$F_2$ — David had access to the system.
$F_3$ — Kyle had access to the system.

Each subfact establishes that the person it names had access. The parent fact r further establishes that the list of access holders is exhaustive. The short answer "Emma, David, and Kyle" expresses the same logical fact r if the question and contract unambiguously establish that the list is exhaustive (T8.4, K4).

Let M be the set of all people who had access to the system under the conditions specified in the question. The content of r entails

$r \vdash (M = \{\text{Emma, David, Kyle}\})$.

Shortening the link text must preserve the exhaustiveness of the list. A fact about granting access does not by itself establish that the person who granted access also held it. If B1 and B8 both meet the conditions of the question, each supplies a separate justification and its own set of possible presentations (K3, K5).

*The graph of the selected justification*

Choose a fact $r \in W$ as the root of the selected analysis, and select the edges $E'$ along which its content is expanded. S contains r and all facts reachable along those edges in the direction from parent to subfact. The path of length zero accounts for r itself:

$$E' \subseteq E_K \cap (W \times W),$$
$$S = \{F \in W : \text{there is a directed path from r to F along edges in } E'\}.$$

The roots of two different analyses may lie in the same W even if neither root preserves the other. Normalization in §4.1 is applied to S for the selected root.

Fully expanding B1 gives all 13 facts in Figure 3.2. At a shallower depth, a composite fact at the boundary of the expansion is retained in full. For example, C1 may be kept whole while C2 is expanded to E1 and C11. E1 then belongs to S along the path through C2 and is also preserved inside C1. A depth of expansion is admissible when the resulting analysis preserves the information, witnesses for relations, and designations needed to answer the question (T9.4, K7).

Before constructing a link, check that r justifies the full content of the answer Â and that the selected analysis contains a fact F designating the same object o as the question and the full answer:

$$r \vdash \hat{A}, \quad F \in S, \quad o \in Obj_T(Q) \cap Obj_T(\hat{A}) \cap Obj_T(F).$$

Such an F belongs to the initial search set. Growing the tree from F covers r and all branches of the selected analysis (§3.4). Since r belongs to S, S entails Â. The presence of r in the contract is checked separately. If the parent fact was derived, the construction takes place in the state where that fact has been registered (K2, K6).

## 4.1. Full Preservation, Absorption, and Normalization S → S′

Consider an analysis with root fact r and fact set S selected in the graph on W (§2.3 and the opening of section 4). Initially each fact in S occupies a separate position. If X fully preserves Y, the content of Y is already represented within X. Absorption removes the separate position of Y while retaining the fact itself, its internal address, and the witnesses for the relations in which it participates (T9.4).

Let $M_i$ be the set of separate positions after i absorptions. To remove Y, exhibit a path of positive length in the original graph S from a preserving fact X to Y. A step removes only the position of Y:

$$M_0 = S,$$
$$X, Y \in M_i, \quad X \succ_S Y \quad \Rightarrow \quad M_{i+1} = M_i \setminus \{Y\}.$$

**Lemma 4.1.1. Preservation of the Justification under Absorption**

*“After each admissible absorption, the content of every original fact and the witnesses for its connections remain accessible in one of the surviving positions.”*

*Proof.* Initially each fact is accessible at its own address. Suppose a step removes position Y and the fact X preserving it remains. For each original fact F previously accessible through Y, extend its internal address by prefixing the path from X to Y. The directions of the connections and the roles of the participants are preserved. In particular, a reference to F continues to point to F inside X. Other addresses remain accessible without modification. Induction proves the claim after every step. □

If S justifies the full content of the answer Â, the remaining positions also justify it: every assertion used from S is preserved in one of them. This applies in particular to the verifiable designations and conditions under which the original assertion was established.

$$S \vdash \hat{A} \quad \Rightarrow \quad M_i \vdash \hat{A}.$$

An absorption sequence is complete when no further position can be removed by this rule. Denote its result by S′. A root of the original graph is a fact with no incoming strict preserving edge. Write Top(S) for the set of these facts.

**Lemma 4.1.2. Independence of Normalization from the Order of Absorption**

*“In a finite nonempty acyclic preserving graph, every complete sequence of absorptions leaves exactly the roots of the original graph.”*

*Proof.* Each step removes one position and leaves another that preserves it. There can therefore be at most $|S| - 1$ steps. A root cannot be removed: the final edge of a path from another fact to that root would enter it.

From any non-root fact Y, ascend against the direction of the edges as long as a preserving fact is available. Finiteness and acyclicity guarantee that the ascent terminates at a root X. In the original graph there is therefore a preserving path from X to Y. Position X survives every step. If position Y also remained at the end, X could absorb it along this path, even if intermediate positions had already been removed (Lemma 4.1.1). This would contradict completeness of the sequence. Thus the result is independent of the order of removal:

$$S' = \mathrm{Top}(S). \ \square$$

This also gives a criterion for an arbitrary finite nonempty acyclic graph. Normalization leaves one position if and only if the graph has a unique root or, equivalently, some fact reaches every other fact:

$$|S'| = 1 \ \Leftrightarrow \ |\mathrm{Top}(S)| = 1 \ \Leftrightarrow \ \exists\, r \in S \ \ \forall\, F \in S: r \succcurlyeq_S F.$$

The first equivalence follows from Lemma 4.1.2. For the second, recall that every ascent from a fact ends at a root. If the root is unique, every fact is reachable from it. Conversely, if r reaches every fact, every fact other than r has an incoming edge, and an edge entering r itself would create a cycle.

In the graph of the selected analysis the unique root is r (§2.3). Hence normalization gives $S' = \{r\}$. In Figure 3.2, the positions of lower-level facts can be absorbed first, followed by the positions of composite facts, while preserving their content in B1. For example, if C2 is absorbed into B1 first, the designation of E1 remains accessible inside B1 at an address through C2. Another address for the same E1, through C1, preserves its participation in a different relation. Ultimately B1 is the only separate position remaining.

(*) In the extended state, absorption likewise requires a path of full preservation. When a fact is a premise of an inference, the record of its role in that inference remains available in the contract. Its position can be absorbed only if a preserving path exists (K6, T9.4).

## 4.2. A Textual Representation of the Root Fact and the Existence of a Single-Link Chain

After normalization, r preserves the entire selected justification. To construct a link, its content must be presented in a text. In the access example the text of a single link presents Emma, David, and Kyle as the exhaustive list of access holders. Each of the three subfacts preserves its participant and that participant's role. The number of links is determined by the number of underlying facts selected, even when several sentences are needed to express one of them (P1, S1).

### Lemma 4.2.1. Presenting an Object in the Text of the Root Fact

*"If a preserving path from r leads to a fact $F_o$ with a verifiable designation of an object o, then r has a full finite representation $V_r$ in which the same object o is designated."*

*Proof.* Choose a path in the graph of the analysis and write it in the direction from parent to subfact:

$$r = F_0 \to F_1 \to \dots \to F_m = F_o.$$

Start with a full representation of $F_o$ that presents a verifiable designation of o (P1). If m = 0, this is already the required text for the root fact.

Suppose that a full text designating o has been obtained for $F_{i+1}$. Use the witness for $F_i \to F_{i+1}$ to select a full representation of the parent $F_i$ in which this occurrence of $F_{i+1}$ is recorded. If the subfact is already presented in full, its text preserves the designations of its objects (P1). For a reference or an abbreviated occurrence, substitute the prepared full text of the subfact at that position. This gives a full finite representation of $F_i$ preserving the parent assertion, the role of the subfact, and the verifiable designation of o (P1–P2). After m steps we obtain $V_r$. □

*Transitions from Question to Link and from Link to Answer*

As at the beginning of section 4, suppose that S is sufficient to justify the full answer and that some fact in S contains a verifiable designation of an object shared by the question and the full answer. Denote this object by o and such a fact by $F_o$. Text construction uses the following properties:

$$S \vdash \hat{A}, \quad F_o \in S,$$
$$o \in Obj_T(Q) \cap Obj_T(\hat{A}) \cap Obj_T(F_o).$$

By §4.1, r entails the entire content of S. The full text $V_r$ expresses r and consequently justifies Â. A preserving path leads from r to $F_o$ (§2.3), so by Lemma 4.2.1 we can choose $V_r$ with a verifiable designation of o. The evidence for resolving references remains available in the same state of the contract. The same object then establishes both boundary connections (P4, S4–S6):

$$p(V_r) \vdash \hat{A},$$
$$Q \leftrightarrow_o (r,V_r) \leftrightarrow_o \hat{A}.$$

It remains to choose a minimally sufficient text for this link. Consider all representations obtainable from $V_r$ by a finite sequence of admissible shortenings with r fixed. Each shortening preserves the justification of the answer, the contribution of the link, and both transitions through o. The initial text also belongs to this family. Denote it by $\mathcal{V}_r$. Since text length is measured in characters, the family contains a representation $V^*$ of least length (P3, P5–P6):

$$V^* \in \mathcal{V}_r, \quad |V^*| = \min \{ |V| : V \in \mathcal{V}_r \}.$$

If $V^*$ admitted a further shortening, the shorter result would belong to the same family. Hence $V^*$ is minimally sufficient in the chain constructed. Take a single link with underlying fact r and this text:

$$Z^* = (r,V^*), \quad L^* = (Z^*),$$
$$Q \leftrightarrow_o Z^* \leftrightarrow_o \hat{A}, \quad p(V^*) \vdash \hat{A}.$$

Thus, for the selected justification with the specified boundary object, an admissible single-link chain exists. Its sole underlying fact justifies the answer, and the designation of o connects the question, the link, and the full answer. □

(*) For a dynamic question the same argument applies after the derived facts have been registered in state $T_t$. The justification in terms of content, the expansion of texts, and both connections through o are checked in that state. The inference records remain available together with the grounds for resolving designations (K6, S6).

## 4.3. Expanding a Link and Choosing the Order of Links

Suppose a link in the constructed chain presents a parent fact P. A more detailed presentation is obtained by selecting several facts nested in P and arranging them in sequence in its place. P itself remains in the graph S and can still be used to verify the relation between its participants. The underlying facts of the new block must be distinct from one another and from the underlying facts of all links retained in the chain (S2). The replacement changes the underlying facts and the link texts selected for the answer.

The question determines how far expansion may proceed. Suppose the parent fact P is expressed by “Emma made dinner and then set the table.” Write A for the fact that she made dinner and B for the fact that she set the table. The question “What two actions did Emma perform?” can be answered by two links, one reporting that she made dinner and the other that she set the table. The parent expresses the relation between the actions, and both texts designate Emma. If the question is “Which action did Emma perform first?”, those two reports taken separately are insufficient. Sentence order in the chain does not by itself assert the order of events. The answer must retain the content of “then” or present other evidence of the sequence (P5, S4–S5).

In Figure 3.2 the parent B1 preserves C1, C2, and B7. Consider replacing one link by three. The relation in the original graph is shown on the left and the proposed order of presentation on the right:

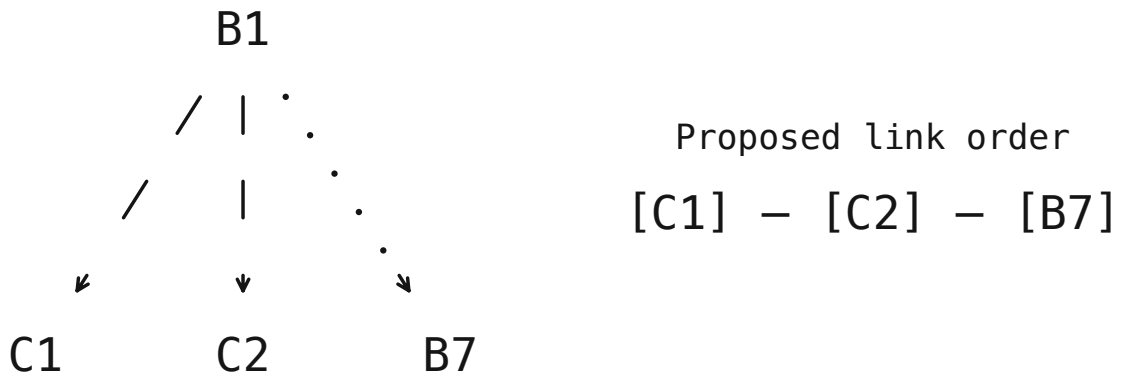

**Figure 4.2.** Replacing link B1 with three nested facts

To establish adjacency of C1 and C2, exhibit the relation of these participants through B1. The shared subfact E1 may contain an object designation for the transition between their texts. B1 also expresses the structural relation between C2 and B7. A designation of a shared object can be sought in E8, nested in both branches. A key is confirmed only if the adjacent link texts actually present designations of the same object (P4, S4).

Let $V_1$, $V_2$, $V_3$ be the texts selected for C1, C2, and B7. In the proposed order we must also check the transition from the question to $V_1$, the transition from $V_3$ to the full answer, and the sufficiency of their joint content:

$$Q \leftrightarrow (C1,V_1) \leftrightarrow (C2,V_2) \leftrightarrow (B7,V_3) \leftrightarrow \hat{A},$$
$$\{p(V_1), p(V_2), p(V_3)\} \vdash \hat{A}.$$

If the required relation is expressed only in B1 and no longer follows from the selected texts, this block cannot replace B1. If every text makes a necessary contribution, all transitions are confirmed, and the justification of the answer is preserved, the block is eligible for further checking as a multi-link presentation.

A verified replacement can be repeated in the resulting chain. At every replacement the conditions are checked for the current texts and adjacencies. By induction, after any finite sequence of accepted replacements the chain justifies the same full answer. Rejecting one replacement does not by itself exclude another order or another admissible text.

Links may be expanded separately. For example, replacing B7 by C10 and C9 gives the order [C1, C2, C10, C9]. The first adjacency is preserved when its texts are unchanged. For C2 and C10 the structural path passes through B1 and B7. For C10 and C9 it passes through B7. E8 and E7, respectively, may carry designations for the corresponding keys. The final boundary now requires a connection between C9 and the full answer. If a link is replaced in the interior of the chain, the transition from the new block to the next link is checked as well. Joint justification and the contribution of each link are assessed for the entire resulting sequence, which uses each underlying fact at most once (S2, S4–S6).

For example, expanding C1 gives subfacts E10 and E1, which are structurally connected through C1. Figure 3.2 does not show whether their texts designate a shared object. Until such a key is exhibited, the replacement of C1 by [E10, E1] remains a candidate.

(*) The same expansion rules apply in a fixed extended state. If the question requests the result of an operation, the joint content of the links must justify that particular result subject to the conditions on the operation (K2, K6).

## 4.4. Checking a Multi-Link Chain and Completeness of the Enumeration of Its Underlying Facts

Once underlying facts and their order have been chosen, the chain must be checked as a whole. Select an admissible text $V_i$ for each underlying fact $F_i$. The same tuple of texts must jointly justify the full answer, preserve the requested relations, and give each link a substantive contribution. Each adjacent pair needs both an expressed structural connection in S and a verifiable lexical key in its texts. Those same texts must also support transitions from the question to the first link and from the last link to the answer. At least one link must contain a confirmed designation of the object common to the question and full answer specified in the theorem (P1, P4–P5, S4–S6).

The simplest case occurs when the question, full answer, and all selected texts designate one object o. It then supplies both boundary connections and every transition between adjacent links. For distinct underlying facts $F_1,\ldots,F_k$ with $k \geq 2$, the connectivity and sufficiency conditions take the form

$$o \in \mathrm{Obj}_T(Q) \cap \mathrm{Obj}_T(\hat{A}) \cap \textstyle\bigcap_i \mathrm{Obj}_T(V_i),$$
$$\mathrm{Rel}_S(F_i,F_{i+1}) \quad \text{for } 1 \leq i < k,$$
$$\{p(V_1),\ldots,p(V_k)\} \vdash \hat{A}.$$

If the texts are admissible, each link's contribution is established, and the evidence for resolving designations is available, these conditions suffice for constructing the chain. In general, successive transitions may use different keys.

For example, in the order [C1, C2, C10, C9], a single representation of C2 must supply both the incoming key with C1 and the outgoing key with C10. If two different shortenings of C2 each work for only one of these transitions, joint compatibility still has to be established. Reversing the order to [C9, C10, C2, C1] also changes the endpoints: C9 must now connect to the question and C1 to the full answer.

Once a compatible tuple of link texts $(V_1,\ldots,V_k)$ has been found, consider all tuples obtained from it by admissible shortenings that preserve the contributions of the links, the justification of the answer, the connection witnesses, and the keys. The initial tuple belongs to this family. Total text length is a nonnegative integer, so the family contains a tuple of least total length. Any further admissible shortening of one text with the others fixed would reduce this sum. Hence each text in such a tuple is minimally sufficient for the chain (P3, P6).

*Enumerating sequences of underlying facts*

Consider the class of presentations that either retain the initial link [r] or replace it by distinct facts nested in r and belonging to S. Different nesting levels and any order without repetition are allowed. In this class r occurs only in the initial single-link chain. Denote the set of facts available for replacement by D. By §2.3, each is reachable from r by a path of positive length.

$$D = S \setminus \{r\}, \quad n = |D|.$$

The link order can interleave facts from different branches. Consider a graph in which P1 and P2 express relations within their respective branches and their common root R connects the branches.

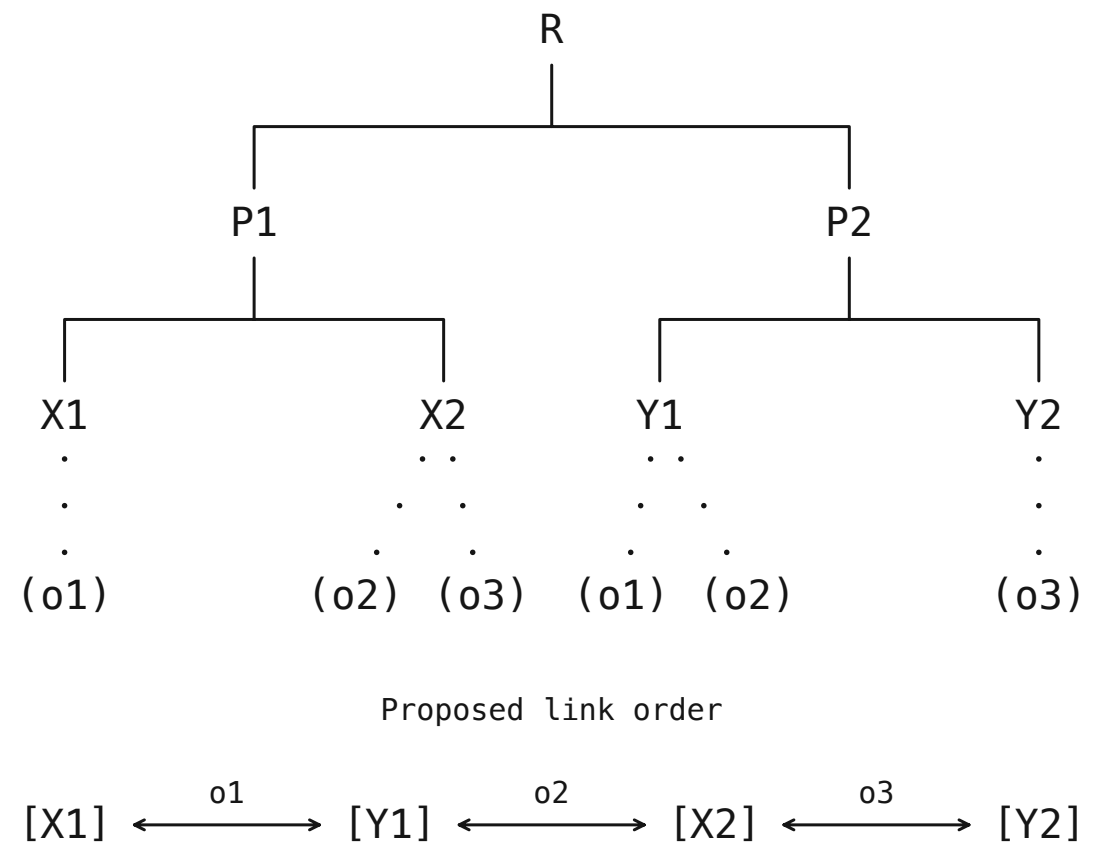

**Figure 4.3.** Interleaving links from two branches

*Notation*

R is the common root of P1 and P2.

X1 and X2 are facts nested in P1.

Y1 and Y2 are facts nested in P2.
Solid lines denote preserving inclusions.
Dotted lines connect a fact to an object it contains.
(o1), (o2), and (o3) are objects. Repeated labels designate the same object.
In the order [X1, Y1, X2, Y2], the structural connections pass through P1, R, and P2. If the selected texts preserve the designations of the objects shown, the internal transitions have the form

$$[X1] \leftrightarrow_{o1} [Y1] \leftrightarrow_{o2} [X2] \leftrightarrow_{o3} [Y2].$$

If each parent were replaced by a contiguous block of its children, X1 and X2 would remain adjacent. Enumeration of individual underlying facts also covers the interleaving shown. The admissibility of this order still depends on the joint content of the texts, the contribution of the links, and both boundary connections.

**Proposition 4.4. Completeness of the Enumeration of Sequences of Underlying Facts**

*"For the finite graph of the selected analysis, enumeration of the initial option [r] and all nonempty sequences of distinct facts from D covers the underlying facts of every admissible chain in the class considered, in their order of occurrence."*

*Proof.* For a sequence of length k, the first fact can be chosen in n ways, the second in n − 1 ways, and so on until the last fact is chosen. Hence the number of sequences of length k and the total number of candidates are

$$N_k = n! / (n - k)!, \quad 1 \le k \le n,$$
$$N = 1 + \sum_{k=1}^{n} N_k.$$

The initial 1 counts [r]. If D is empty this is the only option. In any other chain of the class considered, the underlying facts are distinct and belong to D. The enumeration therefore includes them in exactly the order in which the chain uses them. If an admissible joint presentation exists, the checks above accept that order. Conversely, for every accepted order, such a presentation witnesses the admissibility of the chain. □

This completeness result concerns the underlying facts and their order. Finiteness of S bounds their number, although one fact may have infinitely many admissible texts. Rejecting a single tuple of link texts therefore does not rule out an order of underlying facts. Automatic termination of the selection process would require a separate procedure deciding whether a joint presentation exists. Under the conditions of §4.2, the initial option [r] already has such a presentation. A multi-link option exists only when its texts pass all the checks.

(*) For a dynamic question, S and D are determined in a single fixed state $T_t$. If a new fact is subsequently registered, enumeration in the extended state is relative to the justification selected there.

# 5. A Model for Representing Facts and Expanding Them While Preserving Their Content

In the registry, facts of the contract are represented by finite positive assertions with established participants and relations. These records support both operations used in growing the tree (§3.4): finding the facts preserved by a given record and the facts that contain it. The same record specifies the content to be preserved when constructing a link text (§4.2).

The implementation is restricted to positive assertions with fixed identifiers and unambiguous names. The language has no negation, quantifiers, conditional operators, or modal operators. A mere mention is represented by a separate atom asserting a report or mention, without asserting the content of the target.

## 5.1. Fact Records and Full Content

The registry is a finite sequence of records. Objects and events have stable identifiers. Different mentions of one event retain its identifier, while a new occurrence of the same action receives a different one. Names are associated with participants and events before expansion.

Content is written in atoms, elementary assertions of relations between specified participants. The relation symbol determines what is asserted, and argument positions specify the roles. For example, Before($e_1$,$e_2$) means that event $e_1$ occurred before event $e_2$. The general form is

$$R(a_1,\ldots,a_n).$$

Arguments have specified types: objects, events, or content-based identifiers of previously registered facts. An atom is a constituent of content and does not automatically become a separate fact in the graph. An action and its attributes can belong to a single registered logical unit.

For a record u, let Own(u) denote its own atoms and Inc(u) the previously registered records to which its full inclusions lead. The name of the target of a preserving reference is resolved in the registry. The full content of the record is given by

$$\Sigma_T(u) = \mathrm{Own}(u) \cup \bigcup_{v \in \mathrm{Inc}(u)} \Sigma_T(v).$$

If there are no inclusions, the content of the record is Own(u). The targets have already been registered, so expansion is finite. Every record must have nonempty full content.

Records with the same full content designate one logical fact. Write [u] for the fact represented by record u. Its content $\Sigma_T([u])$ coincides with $\Sigma_T(u)$. Atoms are compared using participant identifiers and argument order. Enumeration order and repetition of the same atom do not affect the result. Records are canonicalized according to

$$[u] = [v] \Leftrightarrow \Sigma_T(u) = \Sigma_T(v).$$

Thus an additional name that simply refers to F does not create another vertex. An abbreviated expression established as a name for F designates its entire content, including the attributes of the event (T8.4, T10.2).

A relation may concern a whole composite fact. Its argument is then a content-based identifier id(F) determined by the full content of the previously registered F. Different names of the same fact yield the same identifier. For example,

$$R(\mathrm{id}(A),\mathrm{id}(B)).$$

When the internal parts of A and B are expanded, the relation continues to refer to the whole facts. Merely mentioning an identifier does not assert the content of its target. Full inclusion of that target is specified separately.

The graph is constructed from the full inclusions and preserving references of all records. If record u contains such an occurrence of record v and the facts [u] and [v] are distinct, the graph contains the edge $[u] \to [v]$. The edge retains the address and witness of every occurrence, including those where several records point to the same pair of facts. If both records designate one fact, the occurrence is recorded within a single vertex. For example, one record may list its atoms explicitly while another presents part of the same content by including a subfact. The inclusion in the second record is available both when expanding the common fact and when ascending from the subfact, whichever name was used to find the common fact.

**Proposition 5.1. Content Preservation and Acyclicity in the Registry**

*"The connections in the registry preserve the full content of facts. Mutual full preservation implies identity, and there are no cycles between distinct facts."*

*Proof.*

The atom set of an included fact is a subset of the union defining its parent's content. For distinct facts this is a strict inclusion, since equal contents correspond to one vertex:

$$P \to F \Rightarrow \Sigma_T(F) \subsetneq \Sigma_T(P).$$

Mutual inclusion gives equality of the sets and identity of the facts. The number of atoms decreases along strict edges, making a return to the starting vertex impossible. □

(*) A derived fact is registered after its justification has been checked. Premises listed in the inference log do not automatically become part of its content. For example, the joint assertion of

the contents of A and B follows from those facts, whereas a new atom asserting causality or succession requires separate justification.

## 5.2. Representations and Content-Preserving Expansion

The full content of a fact is understood as the joint assertion of all its atoms. An interpretation I specifies which relations hold of the fixed participants. Any additional semantic constraints required for the relations are stated explicitly. The registered facts must be jointly satisfiable.

Truth of a fact is defined by

$$I \vDash F \Leftrightarrow \text{for every } a \in \Sigma_T(F),\ I \vDash a.$$

Full inclusion of content therefore ensures entailment: every interpretation satisfying the parent also satisfies all assertions of the nested fact.

$$\Sigma_T(F) \subseteq \Sigma_T(P) \Rightarrow P \vdash F.$$

A representation is constructed from explicit atoms, preserving references, and joint presentations of several parts:

- atom(a) presents atom a.
- $ref^+(f)$ is the full asserted inclusion of the fact unambiguously named by f.
- $and(V_1,\dots,V_k)$ is the joint assertion of the contents of finitely many parts, $k \geq 1$.

Write $C_T$ for the operation that recovers the content of a representation. An explicit atom supplies one assertion, a reference supplies the entire content of its target, and and takes the union of the contents of its parts:

$$C_T(\text{atom}(a)) = \{a\},$$
$$C_T(\text{ref}^+(f)) = \Sigma_T(F), \text{ where ResolveFact}_T(f) = F,$$
$$C_T(\text{and}(V_1,\dots,V_k)) = \textstyle\bigcup_i C_T(V_i).$$

A full representation of F recovers exactly $\Sigma_T(F)$. Its meaning p(V), as used in the main theorem, is the joint assertion of the recovered atoms:

$$C_T(V) = \Sigma_T(F),$$
$$p(V) := \textstyle\bigwedge_{a \in C_T(V)} a.$$

A full representation therefore has the same truth conditions as the fact and preserves its logical consequences. Fact identity itself is determined by equality of the complete sets of atoms, as specified in the canonicalization of records.

For verbal presentation, atoms are assigned unambiguous templates that explicitly name the participants. Reading a template recovers the relation, roles, and identifiers. The and construction is expressed by asserting its parts jointly, possibly in several sentences. A preserving reference designates the inclusion of the named fact in full.

Every fact has a finite explicit representation listing all atoms of $\Sigma_T(F)$ with their participants. When choosing a link text, we check separately that the required object designations are present. Equality of recovered contents alone does not establish this textual condition. Full expansion using the templates presents the required participants directly. Their names are matched to identifiers in the fixed contract (P1, P4).

**Lemma 5.2. Content Preservation under Local Expansion**

*"Replacing a preserving reference by a full representation of its target preserves the content of the entire containing construction and gives a finite representation of the same fact."*

*Proof.*

Select one position for a fully asserted subexpression in a representation. It occurs within an and construction or occupies the entire expression. A participant name, a relation argument, or a mere mention of a fact does not constitute such an expansion position.

Write U[ ] for the construction with the selected position and U[V] for the result of substitution. Let $B_U$ denote the content of the remaining parts, which is independent of the representation inserted.

If the selected position occupies the entire expression, $B_U$ is empty. At each enclosing and construction, the contents of the other parts are added to it. Induction on nesting gives

$$C_T(U[V]) = B_U \cup C_T(V).$$

Let the reference designate fact F, and let $V_F$ be its full representation. The reference and $V_F$ recover the same content. Substituting them in the same construction therefore gives

$$C_T(U[ref^+(f)]) = B_U \cup \Sigma_T(F) = C_T(U[V_F]).$$

All atoms of the parent are thus preserved, including participant roles and relations to whole composite facts. Other reference positions are unchanged, and the fixed resolution of names preserves their targets. Replacing a finite subexpression by a finite admissible expression preserves both finiteness and admissibility of the result. □

For a preserving expansion with the required lexical keys, choose $V_F$ that presents the necessary designations in the part being replaced. The explicit template representation of its atoms permits this choice. Designations outside the replaced position are retained. Content equality and preservation of verifiable designations are therefore established separately. Together they yield the local expansion rule in this implementation (P2). An abbreviated name of a subfact is expanded in the same way.

**Lemma 5.2 allows a designation to be carried along one selected path, as in §4.2. In the positive language of the registry, subfacts whose inclusions are witnessed in different records of the same parent can also be presented jointly (§5.1).**

Let V be a full representation of the parent P. For each selected occurrence, take a full representation $V_i$ of its subfact $F_i$ with the required designations. Adjoin these representations to V using and. Since P fully preserves each $F_i$, we obtain

$$C_T(\text{and}(V,V_1,\ldots,V_k)) = \Sigma_T(P) \cup \bigcup_i \Sigma_T(F_i) = \Sigma_T(P).$$

For example, one record of P refers to A and explicitly states the content of B, while another refers to B and explicitly states the content of A. The full representation and$(V,V_A,V_B)$ has the same content as P. The atoms of V retain the parent's relations and the participants' roles. Each adjoined presentation is associated with the occurrence address and edge witness in the original record. If a subfact occupies two roles, both witnesses are retained.

For the entire selected analysis, apply this construction from the lower facts up to their parents. A fact with no selected subfacts needs only its full representation. Once the subfacts' representations are ready, the equality above gives a full representation of their parent. Since the graph is finite and contains no strict cycles, the construction reaches r. In the resulting text, the selected parts present the verifiable designations, and their participation in parent relations remains verifiable through the witnesses of the original analysis.

(*) Following a valid extension of the registry, the same procedure applies in the fixed state $T_t$. New records refer to previously registered facts. Name resolution and fact content remain fixed during expansion.

## 5.3. Example and Scope of the Implementation

Suppose Kyle went to bed, Emma was cooking, and the child was asleep. The first parent asserts that Kyle went to bed when Emma was cooking. The second asserts that Emma was cooking while the child was asleep.

The records of these facts and their representations take the form

$C1 := \{Event(e_1, Kyle, go_to_bed)\},$
$C2 := \{Event(e_2, Emma, cook)\},$
$C3 := \{Event(e_3, child, sleep)\};$

$V_{B1} = and(ref^+(C1), ref^+(C2), atom(When(e_1,e_2))),$
$V_{B2} = and(ref^+(C2), ref^+(C3), atom(While(e_2,e_3))).$

Here the record $C_i$ specifies the fact's own atoms, and $V_{B1}$ and $V_{B2}$ are representations of the parents. Suppose A1 includes both parents and additionally expresses a specified relation between their contents taken as wholes:

$\rho = JointAccount(id(B1),id(B2)),$
$V_{A1} = and(ref^+(B1), ref^+(B2), atom(\rho)).$

The outer relation ρ is part of the registered assertion. The presence of a shared C2 does not entail it.

```
             A1
            /  \
          /      \
         ↙        ↘
      B1          B2
     /  \        /  \
   /      \    /      \
  ↙        ↘  ↙        ↘
C1          C2          C3
```

**Figure 5.1.** Subfact C2 shared by two parent facts

*Notation*

Arrows denote full preserving inclusions.

C2 is a single fact participating in two parent constructions.

When, While, and ρ are relations preserved together with their participants.

For example, starting at C2, the search discovers both parents, B1 and B2. Once A1 is selected, expansion proceeds downward. Each occurrence of C2 in a parent is preserved at its own address. After A1 has been expanded, three events and all three relations remain:

$$\Sigma_T(A1) = \Sigma_T(C1) \cup \Sigma_T(C2) \cup \Sigma_T(C3)$$
$$\cup \{When(e_1,e_2), While(e_2,e_3), \rho\}.$$

Event $e_2$ is accounted for by a single fact, whose participation in both temporal relations is preserved. The outer relation continues to apply to B1 and B2 as wholes.

To test the merging of repeated presentations, specify a further registry construction. Fact B2 already contains C2. The record A2 additionally specifies a relation σ whose participants are the whole fact B2 and event $e_2$ of C2:

$\sigma = RelatesBlockToEvent(id(B2),e_2),$
$V_{A2} = and(ref^+(C2), ref^+(B2), atom(\sigma)),$
$\Sigma_T(A2) = \Sigma_T(C2) \cup \Sigma_T(C3) \cup \{While(e_2,e_3), \sigma\}.$

Presenting C2 again neither introduces another event nor removes the inner or outer relation. For both constructions, every state of successive reference expansion has been checked programmatically. This finite check pertains to the two records given here.

The search and expansion operations in this example act on logical facts already established. Initial records are selected for an object using verified designations of its identifier. Recognizing such correspondences in unrestricted natural-language text requires separate justification. When constructing a chain, the suitability of the selected underlying fact is checked against the question and the content of the answer under the general conditions of section 4.

(*) In a dynamic application, new facts are first justified and the extended state is fixed. The expansion properties apply to results already registered. These properties of the representation do not establish the correctness of a particular computational operation or guarantee a suitable justification for an arbitrary question.

# 6. Preservation of Results under Changes of Representation

## 6.1. Lexical Reformulation

A single logical fact can be expressed by different lexical constructions. When the wording changes, the graph vertex still represents the same fact with its established content, including events, participants, relations, roles and conditions (T8.4). The number of sentences likewise does not determine the number of vertices. Several sentences can represent one composite parent fact.

Consider two presentations of the same selected justification S. They preserve the logical facts, the directed preserving edges and the witnesses for those edges. If rewording changes the locations of records, a verifiable correspondence is established for the addresses used. Each reference continues to point to the same fact or object in the same role.

Replacing S itself, identifying new logical units or changing the relations between them requires separate consideration. Giving the same answer to one question does not suffice to identify two justifications. For example, a list of two actions and a statement specifying their order can give the same answer to a question about which actions took place, although their full contents differ.

**Corollary 6.1. Invariance of normalization under rewording**

*"Rewording that preserves facts, edges and their witnesses leaves the result of normalization unchanged."*

Let $S'_1$ and $S'_2$ denote the sets remaining after full normalization of the two presentations. Then

$$S'_1 = S'_2 = \mathrm{Top}(S).$$

*Proof.* In both presentations, the same vertices correspond to the same logical facts. The preserving edges retain their directions and witnesses, so the sets of root facts coincide. Full normalization leaves precisely these facts in each presentation (Lemma 4.1.2). For the selected analysis with root fact r, the remaining set in both cases is {r}. □

This comparison concerns the selected S. Comparing the results W of growing the tree also requires matching the initial facts of the search and all incoming and outgoing preserving edges in the reachable region of each presentation. If the initial facts correspond and the correspondence preserves adjacency in both directions, every growth step in one presentation has a counterpart in the other. By §3.4, the results then agree up to this correspondence.

(*) For a dynamic answer, the comparison concerns presentations of one selected justification in a fixed state $T_t$. The same numerical result alone does not establish the identity of facts. The events, the scope and conditions of the operation, and the preserving edges used must also be preserved. Registering new facts that change S goes beyond rewording.

## 6.2. Reproducing a Chain with New Link Texts

When an existing chain is reworded, its selected facts and their order are retained, and the texts of its links are replaced. The replacements must preserve the content that contributes to the answer and the object designations that connect adjacent elements.

Fix a question Q, a full answer Â, an initial justification S and a chain with distinct underlying facts $F_1,\dots,F_k$. Write $V_1,\dots,V_k$ for the original texts and $V'_1,\dots,V'_k$ for the new ones. The replacement takes the form

$$[(F_1,V_1),\dots,(F_k,V_k)] \to [(F_1,V'_1),\dots,(F_k,V'_k)].$$

**Proposition 6.2. Rewording the links**

*"If admissible texts for the same links preserve the content presented by each link and the verifiable designations of the objects used for transitions, the chain with these texts justifies the same full answer."*

*Proof.* The new text of each link preserves the participants, roles and conditions of the original statement. For an abridged text, the comparison concerns the content actually presented in the chain. Thus the links continue to justify the answer jointly, and each retains its contribution (P5, S5).

The underlying facts and their order are unchanged, so the same structural connections are available in S. Check all transition keys using the same tuple of new link texts. Each internal link must present the objects used in both of its transitions. At the boundaries, check the designations in the question and the full answer. Where the original chain required at least one link to designate the selected object shared by question and answer, that designation must also be preserved. If the record addresses change, the available witnesses must still allow the corresponding designations to be resolved (P4, S3–S4, S6).

Let $o_0,\ldots,o_k$ be the objects used in the original transitions. After these checks we obtain

$$Q \leftrightarrow_{o_0} (F_1,V'_1) \leftrightarrow_{o_1} \cdots \leftrightarrow_{o_{k-1}} (F_k,V'_k) \leftrightarrow_{o_k} \hat{A}. \ \square$$

The new wording may admit further abridgment. For these underlying facts in this order, minimally sufficient texts are chosen as in §4.4 from the compatible abridgments of the new tuple of link texts.

(*) In state $T_t$, the new text of a link retains the information and conditions needed to justify the requested result. The premises and the rule used to obtain the result remain available in the inference log. If the chain itself uses the content of a premise, that content is preserved when the text is replaced (K6, S6).

## 6.3. Expanding Repeated Occurrences

A logical fact can participate in several composite structures. Expanding them presents the fact more than once, although a single vertex represents it in the graph. The repeated presentations can be combined while preserving every relation in which the fact participates.

To describe these transformations, write $\langle F_1,\ldots,F_n\rangle$ for a composite fact with specified relations among its participants. The expanded representation $\langle C_1,\ldots,C_m\rangle\mathcal{R}$ presents each distinct nested fact once. $\mathcal{R}$ retains the relations, the roles of the participants, the conditions and their scopes. If a relation applies to a composite fact as a whole, $\mathcal{R}$ also retains its identifier and the grouping of its nested content. The sign ≡ denotes equality of full logical content.

Combining presentations requires them to represent the same full logical fact. Identical wording alone does not establish this. The events, participants and conditions of the statement must coincide (T8.4–T8.5).

*A shared fact in two parent structures*

In the registry example (§5.3), the composite fact A1 expresses a relation between B1 and B2 taken as wholes. The same fact C2 participates in each. Expanding both parents gives

$$A1 = \langle B1,B2\rangle, \quad B1 = \langle C1,C2\rangle, \quad B2 = \langle C2,C3\rangle$$
$$\Rightarrow A1 \equiv \langle C1,C2,C3\rangle\mathcal{R}_1.$$

$\mathcal{R}_1$ includes the relation B1 between C1 and C2, the relation B2 between C2 and C3, and the outer relation A1 between B1 and B2 taken as wholes. A single C2 retains both roles, and the outer relation continues to apply to the original composite facts.

*Direct and nested participation of the same fact*

Consider the test record A2 from §5.3. The path A2 → B2 → C2 already accounts for C2 as part of A2. The direct arrow A2 → C2 in the diagram corresponds to a separately specified role for C2 in relation σ. If the only occurrence of C2 is within B2, the path through B2 suffices to represent it.

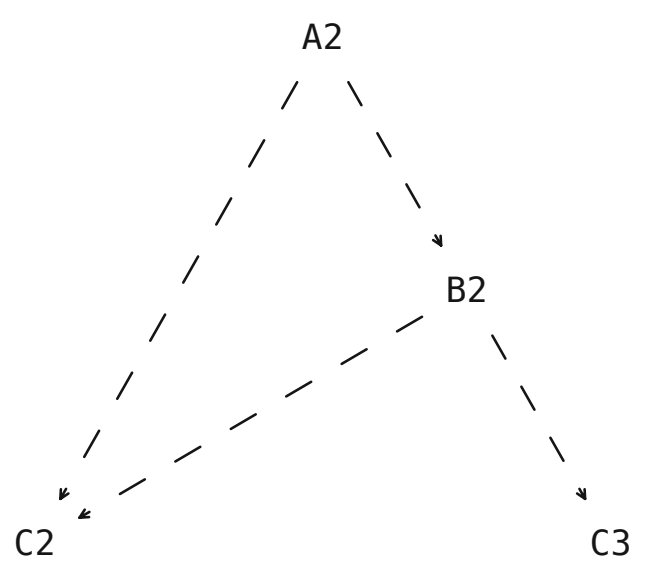

**Figure 6.1.** Test case with two occurrences of C2 in A2

*Notation*

Arrows denote preserving inclusions.

C2 and C3 are nested logical facts.

C2 participates directly in A2 and within B2.

The interior diagonal B2 → C2 shows the inclusion of C2 in B2.

The two presentations of C2 are combined as follows:

$$A2 = \langle C2,B2\rangle, \quad B2 = \langle C2,C3\rangle$$
$$\Rightarrow A2 \equiv \langle C2,C3\rangle\mathcal{R}_2.$$

$\mathcal{R}_2$ retains the inner relation B2 between C2 and C3 and the outer relation A2 between C2 and B2 as a whole. The direct role of C2 in A2 remains distinguishable from its role within B2.

**Proposition 6.3. Combining repeated presentations**

*"Repeated presentations of one logical fact can be combined without changing the full content if all the relations and roles in which that fact participates are preserved."*

*Proof.* Expand the full representations of the nested facts according to P2. In the first case, C2 retains two positions, one in B1 and one in B2. In the second case, it has one position in A2 and one in B2. Assign the identifier of the same C2 to each position. The relation type, the grouping of the participants, their roles and the conditions remain recorded in $\mathcal{R}_1$ or $\mathcal{R}_2$. These records allow both occurrences to be recovered with all their relations. Combining the presentations therefore preserves the full logical content. □

In the first example, a separate relation in A1 connects B1 and B2 as wholes. Their shared C2 does not by itself establish that relation. Combining the two records of C2 retains both positions in which it participates. Normalization, by contrast, removes the independent position of a fact fully preserved by another fact (§4.1).

(*) Two presentations of the derived fact C2 represent the same fact if their full asserted content coincides, including the participants, events, period and conditions of the assertion (T8.4). One fact can have different valid derivations. The premises and rule of each derivation are retained in the log (K6).

## 6.4. Alternative Justifications for the Same Answer

The same full answer can admit several sufficient analyses in the contract. Each analysis is determined by the selected facts and the edges along which its root fact is expanded. Rewording those facts while retaining the edges preserves the analysis (§6.1).

Different chain links can be selected within one analysis. In the example of Emma's two actions (§4.3), A and B are used as links in place of P, while P remains in S as the parent expressing their relation. Removing P from the initial set would require a different justification for the structural connection. Thus selecting the chain [A,B] and replacing S with {A,B} are distinct operations.

Fix a contract T, a question Q and a full answer Â. Consider two analyses of this answer in the region W grown from the objects of the question. Each analysis supplies a root fact $r_i$ expressing the requested relation and specifies the preserving edges along which the facts of $S_i$ are obtained by expanding $r_i$ (§2.3). For each analysis, check sufficiency, the roles of the participants, the inclusion of necessary intermediate facts and the designation of an object shared by the question and the full answer (K5, K7):

$$r_i \in S_i \subseteq W, \quad S_i \vdash \hat{A}, \quad i \in \{1,2\}.$$

Figure 6.2 shows three basic configurations of two analyses. On the left the graphs are disjoint. The middle and right diagrams show two ways in which the graphs can intersect. Distinct root facts can share a subfact, or different branches can expand the same root fact.

```
  No shared facts           Shared subfact            Shared root

R₁          R₂          R₁          R₂                  R
|           |             \        /                   / \
|           |              \      /                   /   \
↓           ↓               ↘   ↙                    ↙     ↘
A           B                 C                     A           B
```

**Figure 6.2.** Configurations of graphs justifying the same answer

*Notation*

$R_1$ and $R_2$ are the root facts of the two analyses in the first two diagrams.

R is the common root fact of the two analyses in the third diagram.

A, B and C are subfacts. Arrows point from parent to subfact.

In the first diagram, consider $R_1 \to A$ and $R_2 \to B$ separately, in the second, $R_1 \to C$ and $R_2 \to C$, and in the third, $R \to A$ and $R \to B$. The same check applies in each case. For each analysis, determine whether its facts and witnessed relations suffice for a full answer to the same question. A shared vertex retains its identity in both analyses.

For example, two statements in the contract answer the question "Did David come into the hall?" One states "David came into the hall before the meeting began" and the other "After David came into the hall, Emma closed the door." In the middle diagram, C is the fact that David came into the hall. Both parents include C, and each independently justifies the answer. In the right diagram there is a single root fact R. If the answer requires the relation between A and B, both branches are expanded together within one analysis.

In each graph, all the selected facts are reachable from its root fact along preserving edges. Hence normalization in the two graphs gives (§2.3, Lemma 4.1.2)

$$\mathrm{Top}(S_1) = \{r_1\}, \quad \mathrm{Top}(S_2) = \{r_2\}.$$

Chains are constructed separately within each original graph (§§4.2–4.4). Justifications for the same answer may contain different additional content, so the root facts $r_1$ and $r_2$ and the admissible orders of the links may differ.

Growing the tree covers all facts reachable from the initial facts (§3.4). Sufficiency has been verified for the analyses presented. How to identify all possible sufficient justifications in W remains an open question.

(*) For a dynamic question, alternative justifications are considered in a single fixed state $T_t$ containing the necessary registered facts. For each result, the premises, the rule applied and the conditions of the inference are available. A premise enters the expansion graph along a preserving edge only if the containing fact fully preserves its content (K6, T9.4).

## Conclusion

A general theory of how information is connected in context must explain how content retains its determinate meaning across utterances. The model proposed here can serve as a foundation for such a theory. Its unit of analysis is the logical fact. Suppose the contract contains a fact that justifies a full answer and that an object shared by the question and the full answer is designated in that fact's expansion along preserving edges. Under the conditions of the model, a search starting from the objects of the question discovers that fact. Its full content can be expanded in a text from which a single link can then be obtained. That link connects the question to the answer and justifies the answer.

The question of what marks the boundary of a logical fact remains open. Utterances may differ in vocabulary, syntax and level of detail yet express the same statement. Conversely, a change in a single designation, the position of a particle or the target of a reference can alter a participant's role or the scope of a condition enough to yield a different fact. The extent of a change in wording does not by itself determine the extent of a change in content. Locating this boundary will require a closer study of the interaction between vocabulary, grammar and the conditions of interpretation.

Here the theory can be developed further. The present model takes the correspondence between a logical fact and its lexical expressions as established. This made it possible to study connectivity among the facts themselves. The next task is to determine what features of a particular language allow us to tell when a change of expression preserves the same logical fact and when it gives rise to a different statement.